\documentclass[sigconf,screen]{acmart}

\AtBeginDocument{%
  }

\setcopyright{none}
\renewcommand\footnotetextcopyrightpermission[1]{}

\makeatletter
\def\@copyrightspace{\relax}
\makeatother

\newcommand\blfootnote[1]{%
  \begingroup
  \renewcommand\thefootnote{}\footnote{#1}%
  \addtocounter{footnote}{-1}%
  \endgroup
}

\acmConference[]{}{}{}
\acmDOI{}
\acmISBN{}
\acmPrice{}

\begin{document}

\title{SAGE: A Strategy-Aware Graph-Enhanced Generation Framework For Online Counseling}


\author{Eliya Naomi Aharon}
\email{eliyaah@post.bgu.ac.il}
\orcid{0009-0009-4641-785X}
\affiliation{%
  \institution{Ben-Gurion University of the Negev}
  \city{Beer Sheva}
  \country{Israel}
}

\author{Meytal Grimland}
\email{meytal.grimland@gmail.com}
\affiliation{%
  \institution{University of Haifa}
  \city{Haifa}
  \country{Israel}}

\author{Avi Segal}
\email{avisegal@gmail.com}
\affiliation{%
  \institution{Ben-Gurion University of the Negev}
  \city{Beer Sheva}
  \country{Israel}}

\author{Loona Ben Dayan}
\email{loona@sahar.org.il}
\author{Inbar Shenfeld}
\email{inbar@sahar.org.il}
\affiliation{%
  \institution{Sahar}
  \city{Tel Aviv}
  \country{Israel}}

\author{Yossi Levi Belz}
\email{yossil@edu.haifa.ac.il}
\affiliation{%
 \institution{University of Haifa}
 \city{Haifa}
 \country{Israel}}
 
\author{Kobi Gal}
\email{kobig@bgu.ac.il}
\affiliation{%
  \institution{Ben-Gurion University of the Negev}
  \city{Beer Sheva}
  \country{Israel}}

\renewcommand{\shortauthors}{Aharon et al.}

\begin{abstract}
Effective mental health counseling is a complex, theory-driven process requiring the simultaneous integration of psychological frameworks, real-time distress signals, and strategic intervention planning. This level of clinical reasoning is critical for safety and therapeutic effectiveness but is often missing in general-purpose Large Language Models (LLMs). 
We introduce SAGE (Strategy-Aware Graph-Enhanced), a novel framework designed to bridge the gap between structured clinical knowledge and generative AI. SAGE constructs a heterogeneous graph that unifies conversational dynamics with a psychologically grounded layer, explicitly anchoring interactions in a theory-driven lexicon.
Our architecture first employs a Next Strategy Classifier to identify the optimal therapeutic intervention. Subsequently, a Graph-Aware Attention mechanism projects graph-derived structural signals into soft prompts, conditioning the LLM to generate responses that maintain clinical depth. Validated through both automated metrics and expert human evaluation, SAGE outperforms baselines in strategy prediction and recommended response quality. By providing actionable intervention recommendations, SAGE serves as a cutting-edge decision-support tool designed to augment human expertise in high-stakes crisis counseling.
\end{abstract}




\maketitle
\blfootnote{\footnotesize \noindent
\copyright\ Eliya Naomi Aharon, Meytal Grimland, Avi Segal, Loona Ben Dayan, Inbar Shenfeld, Yossi Levi Belz, Kobi Gal, 2026. This is the author's version of the work. It is posted here for your personal use. Not for redistribution. The definitive Version of Record will be published in 34th ACM Conference on User Modeling, Adaptation and Personalization (UMAP '26), \url{http://dx.doi.org/10.1145/3774935.3806785}.}

\pagestyle{fancy}
\fancyhf{} 
\renewcommand{\headrulewidth}{0pt} 
\fancyhead[L]{\footnotesize \shorttitle} 
\fancyhead[R]{\footnotesize \shortauthors}

\section{Introduction}
 Successful mental health counseling is an inherently theory-driven process that relies on continuous reasoning about a help-seeker’s evolving emotional and psychological state \cite{gaskell2023effectiveness}. Effective interventions require caregivers to interpret incomplete and often ambiguous signals, integrate them with clinical knowledge, and select therapeutic strategies that are appropriate for the current stage of the interaction.

In online crisis hotlines, which have become commonplace in many countries, these demands are exacerbated under extreme conditions \cite{ni2025scoping}. Caregivers interact with help-seekers through brief, text-only exchanges, where seeker state is only partially observable and may shift rapidly over the course of a conversation. Signals of distress or suicidal risk are often implicit, fragmented across turns, or expressed indirectly. 

\begin{figure}[t] 
  \centering
  \Description{A diagram illustrating a fictitious counseling dialogue divided into three columns: Utterances, Lexicon Category, and Intervention Strategy. The help-seeker's messages are mapped to clinical categories from the SRF lexicon, such as Loneliness, Hopelessness, and Suicidal Ideation. Correspondingly, the caregiver's responses are categorized into intervention strategies like Reflection and Exploration, showing how specific clinical signals trigger distinct strategic interventions.}
  \includegraphics[width=\columnwidth]{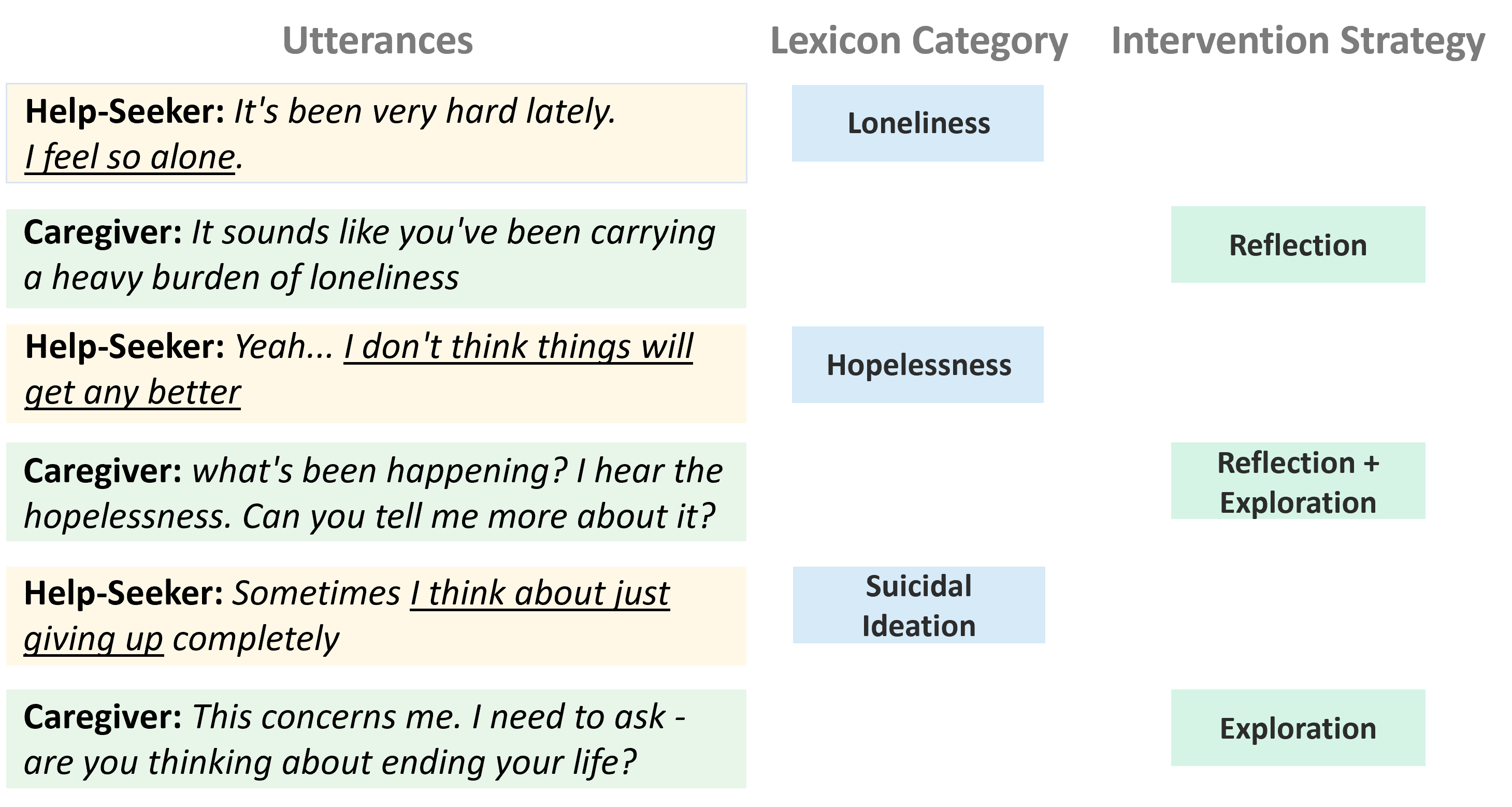} 
  \caption{Fictitious session snippet with psychological categories and intervention strategies presented.}
  \label{fig:diagloue_example}
\end{figure}

Effective support in such settings depends on continuous adaptation: updating internal representations of the help-seeker’s psychological state and selecting intervention strategies that are appropriate. This process requires modeling not only conversational history, but also theory-driven psychological indicators and the strategic intent underlying expert interventions. To illustrate,  Figure~\ref{fig:diagloue_example} presents a fictitious session excerpt between a help-seeker and caregiver with corresponding psychological categories  and intervention strategies.  

At the same time, the growing demand for online crisis support places a substantial cognitive and emotional load on caregivers, who often manage multiple conversations simultaneously and face an increased risk of burnout and  errors \cite{levi2025predicting}. 
Although recent advances in general-purpose large language models (LLMs) have opened new opportunities for AI-assisted mental health support, these models may lack explicit user modeling and clinical grounding, leading to generic or strategically misaligned responses in high-stakes settings \cite{heston2023safety}.

In this work, we introduce SAGE (Strategy-Aware Graph-Enhanced Generation), a framework for decision support in online counseling for recommending responses to caregivers that are aligned with psychological theory. 
SAGE constructs a heterogeneous graph that integrates conversational structure with clinically grounded indicators derived from established suicide-risk frameworks and mental health taxonomies.
It employs an intermediate learning step to infer 
the next intervention strategy to use at each response point in the conversation.
The inferred strategy and graph-derived representations are subsequently used by an LLM to generate a recommended response for the caregiver.

We evaluate SAGE on hundreds of real-world counseling sessions collected from a leading crisis hotline. 
Using a combination of automated metrics and blind human-expert evaluation, we show that SAGE improves both strategy predictions and recommended response quality compared to several baselines, including general purpose LLMs and ablated variants of SAGE.
The advantages of SAGE were apparent even early on in the conversation, where there is limited information about the mental state of the help-seeker.
Our results demonstrate that combining structured representations is 
critical for  effective AI-supported counseling in high-risk online settings. 
More generally, our study highlights the 
advantages of  integrating psychological theory  with user modeling  over general purpose use of LLMs  when supporting mental health.  

\section{Related Work}


Our work intersects three  research strands with relevance to user modeling and adaptation: (1) AI for mental health risk detection, (2) the representation of domain knowledge through graphs and its integration into Large Language Models, (3) strategic reasoning and clinical grounding in AI-driven therapeutic dialogue systems. This section reviews existing literature in these areas and positions SAGE as a framework that bridges the gap between structured clinical protocols and generative capabilities, ensuring that responses are guided by an explicit therapeutic strategy.

A substantial body of work has focused on modeling  mental health risk   from conversational data \cite{bialer2022detecting, izmaylov2023combining}. Subsequent work has enriched these models through structured representations, such as knowledge graphs, to better capture crisis-related signals in professional counseling conversations \cite{xu2021detecting}. 
While these approaches provide increasingly accurate assessments of help-seeker risk, they remain primarily diagnostic, lacking the proactive, intervention oriented recommendations needed for caregiver support. Building on these foundations, SAGE bridges this gap by transforming risk signals into actionable guidance.


Structured representations, particularly graph-based models, have been widely explored in AI for mental health.
By synthesizing structured domain knowledge \cite{welivita2022heal},  frameworks adopt a proactive approach to generate contextually aware responses \cite{deng2023knowledge, deng2025proactive}.
However, many existing systems rely on social media data \cite{welivita2022heal} or conversations simulated by non-clinical crowd-workers \cite{liu-etal-2021-towards}, limiting their alignment with real therapeutic practice. 
A recent review highlights the critical shortage of expert-curated resources and explicitly recommends the development of specialized mental health lexicons to ensure clinical validity \cite{guo2024large}.
Addressing this gap, SAGE constructs its structured representations by bridging expert-led counseling sessions with a theoretical clinical lexicon, thereby facilitating the integration of domain-specific expertise.


Recent studies emphasize the critical role of intermediate reasoning steps, such as dialogue act classification \cite{malhotra2022speaker}, in addressing the limitations of end-to-end generation in mental health contexts.
Empirical evidence demonstrates that without explicit strategic guidance, LLMs exhibit a 'preference bias' toward generic responses regardless of the seeker’s distress \cite{kang2024can}.
To address this limitation, explicitly integrating therapeutic techniques and strategies into prompts improves response quality and alignment with expert behavior \cite{xie2025psydt}.
Complementary approaches use Chain-of-Thought reasoning to first infer psychological and emotional cues before generating a response \cite{wang-etal-2023-cue}.
SAGE extends these works in two ways: First, it incorporates a psychological layer that maps conversational cues onto theoretical clinical categories to infer the help-seeker’s mental state.
Second, it infers the appropriate therapeutic strategy as an intermediate step in the response generation.


Integrating structured external knowledge into LLM-based dialogue systems is an open challenge.
Naively converting knowledge graphs into textual prompts may lead to the loss of structural information and semantics \cite{LLMSurvey}.
Recent work on Graph Neural Prompting demonstrates that naive retrieval can degrade performance due to noise, addressing this issue by using graph neural networks (GNN) encoders to map relational facts into soft prompts that preserve structural integrity  \cite{tian2024graph}.
However, within the dialogue generation task, existing frameworks often revert to converting graph structures into plain text sequences, a process prone to the structural information loss previously noted \cite{tang-etal-2024-cadge-context}.
Furthermore, while recent adaptations for mental health counseling have attempted to integrate external knowledge, they predominantly rely on general commonsense sources to guide sentiment \cite{srivastava2025sentiment},  or extract psychological intentions \cite{peng2022control}, rather than integrating specialized clinical protocols. 
SAGE extends these approaches by including a hierarchical representation of the clinical dialogue in the graph, transforming expert knowledge into strategy-aware soft prompts. This ensures that the structural dynamics are preserved throughout the generation process, aligning the output with expert counseling interventions.

\section{Data Sources}
This section describes the data sources used in this study.
\subsection{Dataset from Online Crisis Hotline}
\label{para:sahar_dataset}
An anonymized  textual conversation dataset was supplied by a national NGO that conducts anonymous  mental-health counseling online. The service is staffed by volunteer caregivers who undergo tailored training to apply established counseling skills in online mental health support. Each session is a one-on-one chat lasting approximately 40 minutes, and the dataset includes the full text of all exchanged messages between a help-seeker and the caregiver supporting them.

At the beginning of each session, help-seekers may provide basic demographic information, specifically age and gender, which is stored as metadata. At the end of the session, caregivers document several key aspects of the interaction, including: the mental-health issues identified (referred to as "distresses"), which encompass 29 categories  (e.g., self-harm, loneliness, anxiety, addiction). 
Additionally, they report the session's effectiveness, which is assessed based on the perceived relief provided to the help-seeker and is rated on a scale from 1 to 5.
All conversations provided by the NGO were tagged as containing suicidal ideation by the caregivers.

\subsection{The SRF Lexicon}
\label{para:srf_lex}

 Our work utilized the Suicide-Risk Factors (SRF) lexicon introduced by ~\citet{grimland2024predicting}, a theory-driven resource designed to capture linguistic indicators of suicide vulnerability. The lexicon is grounded in established psychological frameworks and consists of terms and expressions reflecting both individual and contextual characteristics that are known to be associated with elevated suicidal ideation. This lexicon was initially developed to facilitate suicide-risk prediction by supplying a structured, psychologically informed vocabulary for language-model training. 
 A snippet of a fictitious session and the corresponding categories is shown in Figure~\ref{fig:diagloue_example}.

The SRF lexicon comprises more than 4,000 phrases organized into 20 categories and is arranged hierarchically to integrate multiple complementary psychological models. For example, depressive symptomatology is derived from the Patient Health Questionnaire~\cite{kroenke2001phq}, perceived burdensomeness is informed by the Interpersonal Needs Questionnaire (INQ)~\cite{van2012thwarted}, and indicators of suicidal behavior are based on the Columbia Suicide Severity Rating Scale (C-SSRS)~\cite{posner2008columbia}.

Each category represents a specific dimension of suicide risk. For instance, the hopelessness category includes utterances such as “there is no future for me” or “nothing will ever get better” whereas suicidal intent covers statements like “I have a plan” and “I want to end my life”. By being explicitly grounded in clinical theory and assessment tools, the SRF lexicon supports fine-grained detection and interpretation of diverse suicide-related risk signals in text.


\section{Methodology}

\begin{figure*}
    \centering
    \includegraphics[width=\textwidth]{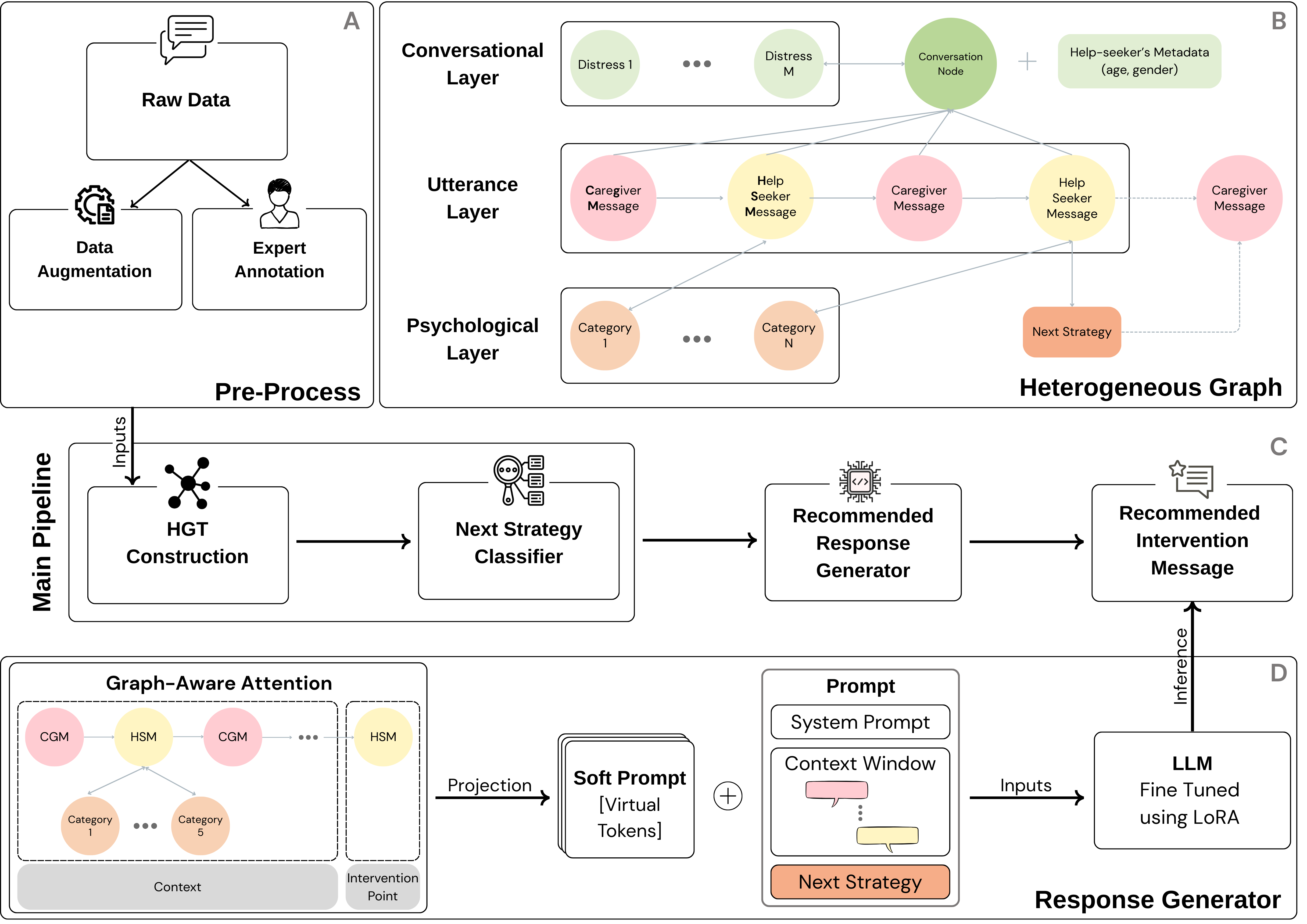}
    \Description{A flow diagram illustrating the SAGE framework, divided into four sections. Section A depicts the pre-processing stage, including data augmentation and expert annotation. Section B provides an illustration of the heterogeneous graph and its three layers. Section C shows the main pipeline, starting with graph construction, followed by the next intervention strategy classification task, and leading to the recommended response generator. Section D details the internal architecture of the response generator, highlighting the graph-aware attention process used to project graph information into soft prompts. These soft prompts are integrated with the conversation context window and the classified intervention strategy into a unified prompt. This combined representation is processed by the LLM, which was fine-tuned end-to-end using LoRA.}
    \caption{Overview of the SAGE framework architecture: (A) Pre-processing involving expert clinical annotation and data augmentation. (B) Heterogeneous Graph Construction comprising three layers: Conversational, Utterance, and Psychological. (C) Main Pipeline featuring a dual-stage task of graph-based Next Strategy Classification followed by strategy-conditioned Response Generation. (D) Response Generator utilizing Graph-Aware Attention to project structural signals into soft prompts, which guide the fine-tuned LLM alongside the predicted strategy and dialogue context.}
    \label{fig:frmework}
\end{figure*}

 In this section we describe each component of SAGE architecture. 
 The main pipeline  (Figure ~\ref{fig:frmework} (C))   employs a dual-stage task at each intervention point: (1) identifying the appropriate intervention strategy for the caregiver, and (2) generating a recommended intervention message that is conditioned on the identified strategy and the learned representation. 
 Further implementation details, including model configurations and hyperparameters, are provided in Section ~\ref{para:implementation_details}.



\subsection{Preprocessing and Strategy Annotation}
\label{para:lexicon_categories}
The pre-processing phase (see Figure ~\ref{fig:frmework} (A)) includes two steps: expert annotation and data augmentation. 
We selected a subset of 150 counseling sessions for expert annotation, prioritizing  the presence of suicidal indicators. 
A licensed clinical psychologist with expertise in suicide detection annotated all caregiver messages in these sessions, categorizing them into one of the three predefined intervention strategies used by the organization: Reflection, Exploration, and Suggestion. 
Drawing from established clinical literature, we define these strategies as follows:
\begin{itemize}
    \item Reflection: A core empathic intervention where thoughts and emotions are echoed and reformulated to convey attentive understanding and provide emotional validation. 
    
    Example: \textit{"I can hear how overwhelmed you feel; however, I also see that you are not giving in to despair, but are instead trying to find ways to ease your burden."}
    
    \item Exploration: An invitational intervention used to prompt the help-seeker to expand upon their narrative or provide additional context. In high-risk scenarios, this includes the explicit assessment of suicidal intent and immediacy to ensure safety. 
    
    Example: \textit{"What you are sharing sounds very concerning. I must ask, are you planning to end your life tonight?"}
    
    \item Suggestion: An intervention aimed at offering new outlooks or coping strategies to encourage cognitive flexibility. This involves highlighting protective factors and reinforcing the impermanence of the current crisis to foster resilience. 
    
    Example: \textit{"You don’t have to go through this alone. Is there someone in your life - a friend, a sibling, or another person you trust - who you could message or call tonight for a bit of support?"}
\end{itemize}
Messages that did not fit these categories, such as greetings, were assigned a Neutral label.  
Based on these annotations, we define the "intervention point" as the last help-seeker message immediately preceding a caregiver message labeled with one of the three intervention strategies, excluding those followed by a neutral label. This process resulted in a total of 2258 annotated intervention points.

In the data augmentation step, the help-seeker utterances are mapped to their corresponding psychological categories of the SRF lexicon described in Section \ref{para:srf_lex}.
This mapping aims to ground the textual input in established psychological theory, transforming raw messages into structured clinical indicators that might be overlooked by raw text alone.
The category mapping was performed by exact string matching, due to the high sensitivity of identifying suicide-related indicators, which requires precision and minimal noise.

\subsection{Heterogeneous Graph Transformer}
\label{para:Graph_Construction}
To capture the multi-layered nature of the counseling sessions, we constructed a  Heterogeneous Graph Transformer (HGT)~\cite{hu2020heterogeneous}.  
It  combines heterogeneous information sources: 
session-level conversational context, structured utterance sequences, and psychological structures grounded in suicide-risk theory.
The HGT consists of three distinct layers, as illustrated in Figure ~\ref{fig:frmework} (B): the conversational layer, the utterance layer, and the psychological layer. Unlike standard GNN that treat all relations uniformly, HGT allows for different types of relationships to be encoded in the graph structure.
The conversational layer captures session-wide information derived from the dataset. 
It contains two types of nodes: distress nodes that represent   the specific mental-health issues documented by the caregiver (see Section ~\ref{para:sahar_dataset}), and conversation nodes that contain  demographic metadata, such as age and gender, providing a profile of the help-seeker. No identifying information is included. 
Within this layer, distress edges connect distress nodes and conversation nodes. 

 The utterance layer consists of message nodes that represent individual utterances from both the help-seeker and the caregiver in a session. Within this layer, temporal edges link messages in chronological order, capturing the sequential structure of the session.
 The psychological layer incorporates the expert-derived categories described in Section ~\ref{para:lexicon_categories}. It consists of lexicon nodes, where each node represents a specific psychological 
factor.

Hierarchical edges in the graph connect message nodes in the utterance layer to their corresponding conversation nodes in the conversational layer, associating individual utterances with session-level attributes. 
Lexicon edges  in the graph connect help-seeker message nodes in the utterance layer to lexicon nodes in the psychological layer, grounding the utterances in clinical indicators to capture the psychological depth of the dialogue.



Table \ref{tab:graph_stats} provides an overview of the HGT components, detailing the distribution of nodes and edges across the three layers. Specifically, the lexicon edges represent the clinical context introduced through the data augmentation process, while temporal edges quantify the sequential dependencies within the sessions.

\begin{table}
\caption{HGT Statistics.}
\label{tab:graph_stats}
\centering
\begin{tabular}{@{}llc@{}}
\toprule
\textbf{Component} & \textbf{Type} & \textbf{Count} \\ \midrule
\textbf{Nodes}     & Utterances & 5,325 \\
                   & Conversations  & 150 \\
                   & Lexicon Categories & 20 \\
                   & Distress Categories & 29 \\ \midrule
\textbf{Edges}     & Temporal & 15,075 \\
                   & Hierarchical & 5,325 \\
                   & Lexicon & 846 \\
                   & Distress & 301 \\ \bottomrule
\end{tabular}
\end{table}

\subsection{Next Strategy Classifier}
\label{para:Next_Strategy_Classifier}
The Next Strategy Classifier  explicitly determines the clinically appropriate mode of intervention before response generation takes place, ensuring that the language model is guided by structured therapeutic strategy.

At each intervention point, the HGT is used to identify the most relevant strategies among Reflection, Exploration, and Suggestion (see Section~\ref{para:lexicon_categories}) to guide the upcoming response.
This task is formulated as a multi-label classification problem where the HGT is trained on annotated intervention points.
It learns dedicated weight matrices for the diverse relation types incorporated in the graph. 
It applies a distinct transformation to each signal based on its relation type, ensuring diverse data sources are processed according to their specific roles.
The resulting HGT representation combines  signals from the preceding utterances, psychological indicators, and help-seeker metadata.

 During inference, the relevant help-seeker node in the utterance layer is passed to a sigmoid-based classification head that predicts the probability of each strategy.
 To handle class imbalance, we calibrated a separate decision threshold for each strategy on a separate validation set. 

\subsection{Recommended Response Generator}
\label{para:Recommended_Response_Generator}

The final stage of the SAGE framework generates a recommended response for the caregiver at each designated intervention point (see  Figure \ref{fig:frmework} (D)). 
This task is performed by a pre-trained LLM and adapted to the clinical domain by employing parameter-efficient fine-tuning using Low-Rank Adaptation (LoRA) \cite{hu2022lora, zheng2023building}.
The framework employs a Graph-Aware Attention mechanism that combines   nodes from the utterance layer and psychological layer of the HGT from the beginning of the session up to the intervention point. 

The learned representation of the help-seeker’s message at the intervention point serves as a query for a cross-attention operation across all previous message nodes in the utterance layer and their associated  nodes in the psychological layer. The resulting representation is  passed through a domain projector, which produces a soft prompt that is inputted into the LLM.  The final generation prompt integrates three primary inputs: (1) the soft prompt containing learned structural representations, (2) the next intervention strategy to implement (as identified by the Next Strategy Classifier), along with its clinical definition, and (3) a sliding window of the most recent messages.


\section{Experimental Setup}
In this section we describe the experimental setup for evaluating SAGE. Our evaluation assesses separately the framework’s performance for each task (predicting the next intervention strategy, and generating a recommended response)  at each designated intervention point. For both tasks we compare the performance of SAGE   
 against several ablated configurations and baselines. 
To further validate the framework's performance,  a human expert performed a blind pairwise comparison between  the response generated by SAGE and a baseline.

\subsection{Implementation Details}
\label{para:implementation_details}


All architectural decisions and training hyperparameters were empirically tuned to improve performance and ensure generalization.
Implementing the methodology (~\ref{para:Graph_Construction},~\ref{para:Next_Strategy_Classifier}), the classifier's experimental settings are as follows: the message nodes were initialized using the CLS representation from AlephBERT \cite{seker2022alephbert} ($Dim=768$), a transformer model pretrained on Hebrew, adopting a content-based initialization shown to outperform random approaches \cite{spillo2024evaluating}.
These representations were augmented with a positional feature normalized to $[0, 1]$ relative to session length, indicating each utterance's location.
The lexicon and distress nodes were represented by trainable embeddings with a dimension of 32. 
These embeddings were optimized during training.
Distress nodes were masked during inference to ensure a realistic setting.
The HGT architecture utilized a hidden dimension of 256 across 2 layers, with 4 attention heads per layer and a dropout rate of 0.2. 
For optimization, we employed the Adam optimizer with a learning rate of $1 \times 10^{-3}$ and a weight decay of $1 \times 10^{-5}$. 


Training utilized a weighted binary cross-entropy loss to address class imbalance. 
Maximizing validation Macro-$F_1$ performance guided both the selection of the best performing model weights and the calibration of thresholds for each intervention strategy.


We used the Gemma-3-12b-it \cite{gemma_2025} LLM to generate the recommended response. The data obtained from online crisis hotline was in Hebrew, and this model was selected for its multilingual proficiency and suitability for local deployment, a setup mandated by our dataset sensitivity. 

To adapt the response generator, we applied LoRA with a rank ($r$) of 16, an alpha ($\alpha$) of 32, and a dropout rate of 0.1. The LoRA layers were integrated into all linear modules within the transformer blocks, including both the attention and MLP layers.

Consistent with the Graph-Aware Attention mechanism (Section ~\ref{para:Recommended_Response_Generator}), structural context is integrated via multi-token soft prompting, which maps graph-derived signals into a sequence of continuous virtual tokens. 
This mechanism enables the model to selectively attend to structural signals across the entire subgraph, capturing long-range dependencies that may exceed the textual prompt's capacity. 
The resulting context is then projected into 8 virtual tokens that precede the text embeddings in the LLM’s input sequence.
A learned gating parameter, initialized at 0.5, dynamically regulates the influence of these graph-derived tokens relative to the textual input. 
To direct the generation according to the predicted intervention, the input prompt includes the next strategy to implement, alongside a sliding window of the 5 most recent utterances.

We used the AdamW optimizer with distinct learning rates: $5 \times 10^{-6}$ for LoRA adapters and $5 \times 10^{-2}$ for graph projection and gating parameters. 
Training proceeded for up to 7 epochs and was monitored via validation loss, with early stopping after 2 epochs to support generalization.

\section{Results} 

Building on the 150 expert-annotated sessions and 2258 intervention points established in Section ~\ref{para:lexicon_categories}, we partitioned the data into training (75\%), validation (12.5\%), and test (12.5\%) sets. To ensure reproducibility, the partition was executed at the conversation level using a fixed random seed (seed=42), preventing data leakage by ensuring all messages from a single session remain within the same split.  
We report results separately for the two integrated learning tasks on the test set.

\subsection{Next Strategy Classification}
\label{para:result_NIS}
 We evaluate the full HGT model against variants that systematically remove specific layers to isolate their impact on performance.  
The evaluation metrics included  F1 \cite{fu2025ham,deng2023knowledge,malhotra2022speaker}, and the Matthews Correlation Coefficient (MCC) to account for the class imbalance within the intervention strategies.

Table \ref{tab:ablation} compares the performance of the Next Strategy Classifier for three configurations: A flat model using a  BERT model (AlephBert),    an ablation variant of SAGE that   only included  the utterance layer in the HGT architecture, and the full SAGE model. 
As shown in the Table, while the baseline AlephBERT model achieved a limited MCC of 0.12, we observed a consistent improvement using the graph-based frameworks, with the full SAGE model achieving best performance (MCC of 0.59). The transition from  the flat model  to the Utterance layer  configuration yielded the most substantial performance jump, from MCC of 0.12 to 0.57. 
A similar trend was observed for the F1 metric.
These results  align with other works showing that   graph-based architectures  provide a more robust representation of dialogue dynamics than sequential models~\cite{fu2025ham}. 
The performance gains observed upon the integration of the Conversation and Psychological layers in the HGT  indicate that broader context and clinical signals are essential for best performance.


\begin{table}
\caption{Performance comparison of Next Strategy Classification across conditions.}
\label{tab:ablation}
\centering
\small
\begin{tabular}{lcc}
\hline
\textbf{Model} & \textbf{F1} & \textbf{MCC} \\
\hline
AlephBERT (No Graph) & 0.60& 0.12 \\
Utterance layer only & 0.76  & 0.57 \\
Full HGT (SAGE) & \textbf{0.77} & \textbf{0.59} \\
\hline
\end{tabular}
\end{table}

Table \ref{tab:per_label} breaks down the performance of SAGE for predicting the different strategies.
As shown in the Table, results demonstrate high proficiency in identifying Exploration ($F_1=0.84$) and Reflection ($F_1=0.77$). In contrast, the performance for Suggestion is lower ($F_1=0.65$).
This gap is driven by two primary factors.
First, Suggestion is less frequent and rarely occurs in isolation.
While Reflection and Exploration appear frequently as standalone strategies (30.82\% and 27.59\%  respectively), Suggestion is less common (12.58\%) and more likely to occur in combination with other strategies (14.9\%).
Second, Suggestion is inherently more complex.
Reflection and Exploration act as immediate responses to local cues and distress markers. They typically address the help-seeker's most recent utterance, making them easier to model compared to Suggestion, which requires synthesizing the evolving conversational context to assess the seeker's receptivity to actionable advice based on a personalized understanding of their specific needs.

\begin{table}[h]
\caption{Per-label performance for Next Strategy Classification.}
\label{tab:per_label}
\centering
\small
\begin{tabular}{lccccc}
\hline
\textbf{Strategy} & \textbf{Precision} & \textbf{Recall} & \textbf{F1} & \textbf{MCC} & \textbf{Num.} \\
\hline
Reflection & 0.69 & 0.87 & 0.77 & 0.58 & 123 \\
Exploration & 0.78 & 0.92 & 0.84 & 0.67 & 144 \\
Suggestion & 0.73 & 0.58 & 0.65 & 0.50 & 100 \\
\hline
\end{tabular}
\end{table}

\subsection{Recommended Response Generation}
\label{para:response_results}
 For response generation, we compare the full SAGE configuration against several baseline settings to evaluate the influence of graph-aware soft prompting and intervention strategy prediction.
We define four experimental configurations:
\begin{itemize}
    \item Vanilla (baseline): A pre-trained Gemma model using only the dialogue context window as input.
    \item Vanilla FT: An ablated version of SAGE using only the dialogue context window.
    \item Graph Aware (GA FT): An ablated version of SAGE that incorporates the graph structure and soft prompts but excludes strategy prediction.
    \item SAGE: The complete SAGE framework.
\end{itemize}
The evaluation metrics included the BERTScore \cite{zhang2019bertscore,xie2025psydt,srivastava2025sentiment},  which measures the  semantic similarity between generated responses and human-caregiver gold standards. Additionally, Perplexity was used to measure the linguistic fluency and coherence of the model output \cite{deng2023knowledge,tang-etal-2024-cadge-context}.

Table \ref{tab:generation_comparison} details the performance of the various model configurations across automatic evaluation metrics. The results demonstrate a consistent progression in generation quality as SAGE components are integrated.
Fine-tuning yields clear improvements in both semantic similarity to the gold standard and fluency.
Notably, the GA FT configuration demonstrates that integrating graph-aware attention via soft prompting not only enhances semantic alignment but also remarkably improves model fluency.
Finally, the full SAGE configuration achieves the highest BERTScore (0.701), representing the strongest alignment with expert messages.

We used a Friedman test as the non-parametric omnibus test, followed by pairwise Wilcoxon signed-rank tests with Holm correction ($\alpha = 0.05$, two-sided).
For BERTScore, we found a significant effect of condition (Friedman $\chi^2 = 272.02$, $p < 0.001$). Post-hoc tests revealed that SAGE significantly outperformed both Vanilla ($p < 0.001$) and Vanilla FT ($p < 0.001$), while the difference between SAGE and GA FT was not significant. 

Perplexity also showed a significant effect of condition (Friedman $\chi^2 = 154.44$, $p < 0.001$). Interestingly, GA FT achieved lower perplexity than SAGE ($p < 0.001$), suggesting that explicit strategy conditioning may shift the model toward more structured interventions that prioritize semantic alignment over lexical similarity with historical responses. Both graph-aware configurations substantially outperformed the baseline models (all $p < 0.001$).

Overall, these results indicate that incorporating graph-aware representations capturing conversational context and clinical signals leads to responses that better align with therapeutic intent, as reflected by both BERTScore and perplexity.

\begin{table}[h]
\caption{Comparison of generation configurations across automatic metrics.}
\label{tab:generation_comparison}
\centering
\begin{tabular}{lccc}
\hline
\textbf{Configuration} & \textbf{BERTScore} & \textbf{PPL}  \\
\hline
Vanilla          & 0.617 & 540.488  \\
Vanilla FT                & 0.690 & 182.255  \\
GA FT        & 0.698 & \textbf{12.709}  \\
SAGE   & \textbf{0.701} & 26.153 \\
\hline
\end{tabular}
\end{table}

\subsection{Human Evaluation}
To further evaluate the clinical validity and practical utility of SAGE, we conducted a blind pairwise comparison against the Vanilla FT for 307 samples. An expert evaluator (licensed clinical psychologist with expertise in suicide ideation prevention) was provided with the full session history up to each intervention point and presented with the generated responses from two models: SAGE and Vanilla FT. The evaluator was asked to select their preferred response or indicate a tie. 
The expert was also requested to justify each choice by selecting one or more of the following criteria:
\begin{itemize}
    \item Empathy: The response felt more human and supportive.
    \item Relevance: The response was more contextually relevant to the specific session history.
    \item Safety: The response was safer and handled risk better.
    \item Fluency: The response was easier to read, had better grammar, and avoided repetition.
\end{itemize}


The benefit  of such comparative approaches was demonstrated in past work \cite{van2021human}, with evaluation criteria grounded in established mental health and response generation literature \cite{peng2022control, althoff2016large,xie2025psydt,deng2023knowledge}.

To ensure the reliability of the subjective assessments, we implemented a quality control measure by including a subset of repeated instances. This allowed us to calculate intra-rater consistency, verifying that the evaluator's judgments remained stable throughout the evaluation session.
The analysis yielded an intra-rater consistency score of $\kappa = 0.91$ (Cohen's Kappa), indicating almost perfect agreement. This high level of reliability confirms that the clinical distinctions captured in the evaluation are stable and grounded in their professional expertise.

\begin{figure}[h]
    \centering
    \Description{A pie chart illustrating the distribution of expert preferences across 307 samples. The chart shows that SAGE was preferred in 50.2\% of the cases, Vanilla FT was preferred in 34.5\% of the cases, and the remaining 15.3\% were judged as ties.}
    \includegraphics[width=0.6\columnwidth]{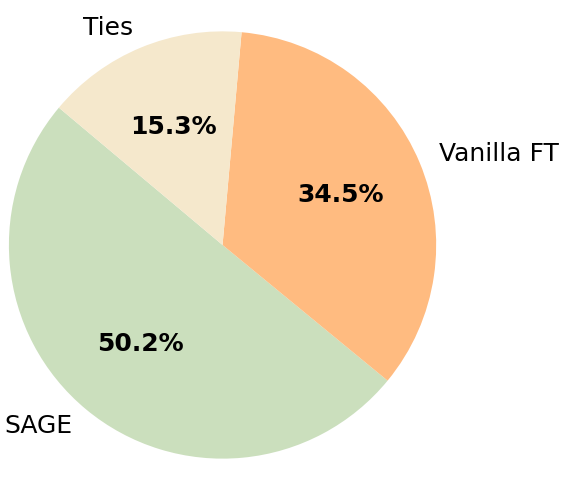}
    \caption{Distribution of expert preferences over SAGE and Vanilla FT.}
    \label{fig:pie-chart}
\end{figure}

Figure \ref{fig:pie-chart} illustrates the distribution of the expert preferences across the 307 samples.
As shown, there were significantly more cases in which the 
expert preferred SAGE to the Vanilla FT. 
Specifically, the expert evaluator preferred SAGE in 50.2\% of the cases, while Vanilla FT was preferred in 34.5\%. In the remaining 15.3\% of instances, the models were judged to be equivalent in quality.
A binomial test against the null hypothesis $p=0.5$ confirmed that the preference of SAGE was statistically significant ($p \approx 0.006$), with a 95\% confidence interval for the win probability of approximately [0.52, 0.64]. This indicates expert preference for SAGE, while accounting for the substantial proportion of ties.

To better understand the qualitative drivers behind the expert's opinions, we analyzed the specific criteria cited by the evaluator in cases where SAGE was preferred. Our analysis reveals that in 97.4\% of the cases where SAGE was preferred, expert justification included  empathy, reflecting a more supportive and human-like tone. Similarly,   in 77.92\% of these instances, SAGE's responses were marked as more contextually relevant. Fluency and Safety were also identified as deciding factors in 63.64\% and 37.01\% of SAGE’s winning cases, respectively.
Furthermore, all four clinical criteria were simultaneously cited in 26.6\% of SAGE’s winning cases, compared to only 18.9\% of instances where the Vanilla FT was preferred.



\begin{figure}[h] 
    \centering
    \includegraphics[width=\columnwidth]{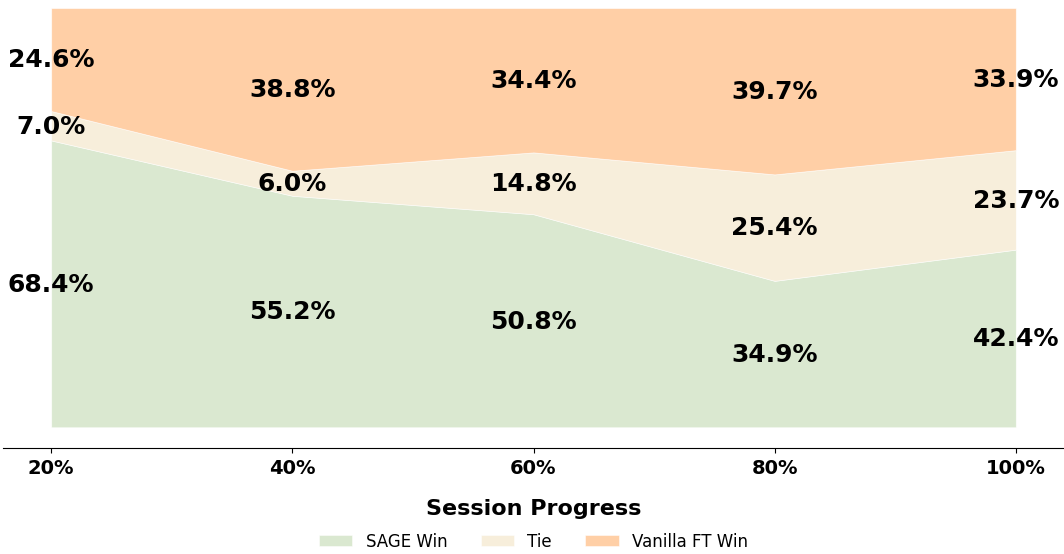}
    \Description{A stacked area chart illustrating the flow of expert preferences across session progress, divided into 20\% intervals. The dark green section (SAGE Win) starts with a strong lead of 68.4\% at the 20\% mark and remains above 50\% through the 60\% progress interval. The light beige section (Vanilla Win) fluctuates between 24.6\% and 39.7\%. The middle beige section representing ties increases significantly from 7.0\% at the beginning to 23.7\% at the session's conclusion (100\% mark).}
    \caption{Analysis of  expert preferences across session stages.}
    \label{fig:preference_flow}
\end{figure}

We further analyze the human expert evaluations of both conditions by examining how preference patterns change as a function of the intervention’s position within the session (Figure \ref{fig:preference_flow}). To allow comparison across sessions of different lengths, each dialogue is normalized into relative percentage intervals, reflecting the progression of the counseling process from its beginning to its end.
As shown in the figure,  SAGE demonstrates a clear advantage over the Vanilla FT baseline, particularly in the earlier stages of the interaction, where strategic guidance and psychological grounding are most critical for shaping the trajectory of the session. 
This performance is noteworthy given that  information about the help-seeker's mental state  and the appropriate strategy is typically limited at the beginning of a session.
While the preference rate for the Vanilla FT remains relatively stable across the sessions, the proportion of ties increases toward later stages where more information is available.

\begin{table*}[h]
\footnotesize
\centering
\caption{Qualitative comparison of model responses in representative use cases.}
\label{tab:use_case_comparison}
\begin{tabular}{p{0.01\textwidth} p{0.25\textwidth} p{0.21\textwidth} p{0.21\textwidth} p{0.21\textwidth}}
\hline
\textbf{\#} & \textbf{Context Summary} & \textbf{Last help-seeker message} & \textbf{Vanilla FT} & \textbf{SAGE} \\ 
\hline
1 & The help-seeker is in a crisis state, recovering from a recent manic episode. She reports significant family conflicts and mentions being home alone, experiencing suicidal ideation. & \textit{"What should we do so that I don't do something I'll regret?"} & I hear how worried you are that you might do something you don't want to... maybe it's worth trying to go to sleep now. & Let's try to think together about what can be done to ease this pain. \\ 
\hline
2 & The help-seeker describes exhaustion and a total lack of energy, stating that every day feels like a 'war'. He shares the pain of a broken heart after opening up to someone and being rejected. & \textit{"I ask myself every day what I’ve done in my life to deserve this. I take blow after blow, one bad thing after another. I have no strength left; sometimes I just want to end it."} & Dear, I hear how broken your heart is... how much you are suffering... I hear how lost you feel... I am worried about you... & Dear, I hear the despair... the pain... the urge to stop... I am worried about you... Do you have an intention to die tonight? \\ 
\hline
\end{tabular}
\end{table*}

\section{Discussion}

Our results demonstrate that explicit modeling of help-seeker psychological state and intervention strategy improves both strategy prediction and response quality in crisis counseling support. The performance gain from text-only baselines to graph-based architectures (MCC increasing from 0.12 to 0.57) underscores the value of structured representations in capturing dialogue dynamics, extending prior findings on GNN superiority in dialogue tasks \cite{fu2025ham} to the specialized domain of crisis counseling. The additional integration of the psychological layer provided more modest but meaningful improvements (MCC from 0.57 to 0.59), suggesting that theory-driven clinical indicators offer complementary information that refines model decisions when explicit markers are present.

Human evaluation provides compelling evidence for SAGE's practical utility, with the expert evaluator preferring SAGE responses in 50.2\% of cases, primarily driven by greater empathy (97.4\%) and contextual relevance (77.9\%). The temporal analysis (Figure ~\ref{fig:preference_flow}) reveals that SAGE's advantage is most pronounced in earlier session stages,
suggesting that SAGE effectively infers psychological states from the limited initial context, enabling it to proactively guide session trajectories from the outset.

The variations in performance across intervention strategy types (Reflection F1=0.77, Exploration F1=0.84, Suggestion F1=0.65) reflect varying complexity: while Reflection and Exploration primarily leverage local cues within recent utterances, Suggestion requires synthesizing context from the entire session to assess readiness for action-oriented guidance. 
Furthermore, the perplexity-BERTScore trade-off, where SAGE achieved the highest semantic alignment (0.701), despite GA FT demonstrating lower perplexity, suggests that explicit strategy conditioning promotes clinically principled responses that may diverge from exact training phrasing, prioritizing therapeutic intent over lexical mimicry.


To contextualize the quantitative findings, we analyze representative use cases that demonstrate the practical advantages of SAGE’s architecture in shaping session trajectories (Table \ref{tab:use_case_comparison}).
In Use Case 1, while the Vanilla FT baseline offers a standard reflection, its subsequent suggestion (\textit{'...go to sleep now'}) preemptively closes the dialogue, leaving the help-seeker alone with suicidal ideation. 
Conversely, SAGE’s collaborative 'think together' approach fosters engagement and sustains the intended intervention strategy.
In Use Case 2, while Vanilla FT provides empathy, it fails to prioritize the help-seeker's critical distress signal (\textit{'I just want to end it'}).
In contrast, SAGE validates the help-seeker's pain and performs a direct risk assessment by explicitly inquiring about suicidal intent, thereby demonstrating adherence to clinical safety protocols.


\subsection{Limitations}
Several limitations  of this study warrant consideration. 
While the requirement for expert-level annotation ensures high clinical validity, it inherently creates a resource bottleneck that limits overall dataset scale. This fact may also constrain broader generalizability across diverse counseling contexts. 
Furthermore, since our evaluation is conducted at discrete intervention points within static dialogue logs, it does not capture the downstream  influence of SAGE's recommendations. Moving beyond this would require the implementation of interactive simulators or human-in-the-loop frameworks to fully evaluate how SAGE’s strategy-driven recommendations actively shape session trajectories in real-time.

\section{Conclusion}

Effective mental health counseling requires simultaneous integration of psychological frameworks, real-time distress signal identification, and strategic intervention planning - a level of depth often missing in general-purpose language models. This work introduced SAGE, a framework that bridges structured clinical knowledge with generative AI by constructing heterogeneous graphs unifying conversational dynamics with theory-driven psychological indicators.

SAGE's architecture employs a two-stage approach: first identifying the appropriate intervention strategy through graph-based classification, then conditioning response generation on both this predicted strategy and learned structural representations via soft prompting. Validated on real-world crisis hotline sessions through automated metrics and blind expert evaluation, SAGE demonstrates that explicit modeling of user state and proactive strategic intent may improve the clinical relevance and empathy of AI-generated support.

The framework's architectural principles: heterogeneous graph construction grounding dialogue in domain theory, structure-preserving soft prompting, and staged inference separating state estimation from strategic adaptation - offer transferable approaches for adaptive systems in specialized domains requiring nuanced user modeling and strategic responsiveness.

Future work will integrate expert feedback through human-in-the-loop learning, utilize larger and more diverse datasets, conduct longitudinal deployment studies measuring impact on caregivers’ decisions and help-seeker outcomes, and extend to multilingual contexts. As online mental health demand grows, AI must evolve beyond generic capabilities toward specialized tools embodying clinical expertise. SAGE represents this direction: maintaining clinical depth while providing scalable decision support for caregivers navigating high-stakes crisis intervention.
\paragraph{Ethical Considerations} This research was approved by the Institutional Review Board (IRB). Both help-seekers and caregivers provided informed consent for the use of anonymized data for research purposes.

\bibliographystyle{ACM-Reference-Format}
\bibliography{references}










\end{document}